\documentclass[letterpaper,10pt,conference]{ieeeconf}  % 

\makeatletter
    \let\NAT@parse\undefined
\makeatother

\usepackage{url}
\usepackage{multicol}
\usepackage[T1]{fontenc} % to get rid of some font warning
\usepackage{graphicx}
\usepackage{float}
\usepackage{mathtools, algorithm, algorithmicx}
\usepackage[font=small, labelfont=bf]{caption}
\usepackage{subcaption}
\usepackage{listings}
\usepackage{algpseudocode}
\usepackage{multirow}
\usepackage[tikz]{mdframed}
\mdfsetup{innertopmargin=1pt} 
\usepackage{tabularx}
\usepackage{booktabs} % For better rules
\usepackage{array} % For better column definitions
\usepackage{url}
\usepackage{lipsum}

\usepackage[bookmarks=true,pagebackref=true,colorlinks=true]{hyperref}
\usepackage{cleveref}
\usepackage{amsmath,amssymb,array,balance,booktabs,cite,changes,color,enumerate,float,graphicx,hyperref,multicol,multirow,siunitx,times,subcaption,mathtools,colortbl}
\usepackage{wrapfig}
\usepackage[export]{adjustbox}
\usepackage{xcolor}
\usepackage[font=footnotesize]{caption}
\usepackage[export]{adjustbox}
\usepackage{algorithm}
\usepackage{xspace}
\definecolor{myblue}{rgb}{0.44, 0.65, 0.82}
\definecolor{myred}{rgb}{0.82, 0.65, 0.44}
\definecolor{myorange}{rgb}{0.82, 0.4, 0.1}
\definecolor{mypink}{rgb}{0.98, 0.671, 0.843}
\definecolor{mygreen}{rgb}{0.032, 0.6392, 0.2039}
\newcommand{\videourl}{\url{https://lauramsmith.github.io/steer}}

\newcommand{\alert}[1]{\textbf{}}

\newcommand{\BEAS}{\begin{eqnarray*}}
\newcommand{\EEAS}{\end{eqnarray*}}
\newcommand{\BEA}{\begin{eqnarray}}
\newcommand{\EEA}{\end{eqnarray}}
\newcommand{\BEQ}{\begin{equation}}
\newcommand{\EEQ}{\end{equation}}
\newcommand{\BIT}{\begin{itemize}}
\newcommand{\EIT}{\end{itemize}}
\newcommand{\BNUM}{\begin{enumerate}}
\newcommand{\ENUM}{\end{enumerate}}
\newcommand{\BEL}[1]{\begin{equation}\label{#1}}
\newcommand{\EEL}{\end{equation}}
\newcommand{\BA}{\begin{array}}
\newcommand{\EA}{\end{array}}

\hypersetup{colorlinks = true, breaklinks = true, citecolor = [rgb]{0,0.07843,0.45098}, urlcolor = [rgb]{0,0.07843,0.45098}, linkcolor = [rgb]{0,0.07843,0.45098}} 

\newcommand{\metabbr}{STEER\xspace}

\usepackage{inconsolata}
\usepackage[T1]{fontenc}

\definecolor{light-gray}{rgb}{0.8, 0.8, 0.8}
\definecolor{comment-green}{rgb}{0.435, 0.576, 0.106}
\definecolor{prompt-blue}{HTML}{2596be}
\definecolor{code-function}{HTML}{379fbe}
\definecolor{code-function}{HTML}{693da8}  % brian (maybe remove)
\definecolor{code-syntax}{HTML}{0060b1}
\definecolor{code-constant}{HTML}{d86001}
\definecolor{prompt-gray}{HTML}{a7a7a7}
\definecolor{highlight}{HTML}{f8f9cb}
\definecolor{highlight}{HTML}{e3eeff} 
\definecolor{code-perception}{HTML}{2ecc71}
\definecolor{code-control}{HTML}{ff9900}
\definecolor{code-undefined}{HTML}{ff0000}
\renewcommand\fbox{\fcolorbox{light-gray}{white}}

\NewDocumentCommand{\code}{v}{%
\texttt{\small{\textcolor{code-function}{#1}}}%
}
\usepackage{selectp}

\title{\LARGE \bf STEER: Flexible Robotic Manipulation via Dense Language Grounding}

\author{
\authorblockN{Laura Smith\textsuperscript{1,2}, Alex Irpan\textsuperscript{1}, Montserrat Gonzalez Arenas\textsuperscript{1}, Sean Kirmani\textsuperscript{1},
\\
Dmitry Kalashnikov\textsuperscript{1}, Dhruv Shah\textsuperscript{1}, Ted Xiao\textsuperscript{1}}
\authorblockA{\textsuperscript{1}Google DeepMind \,\, \textsuperscript{2}UC Berkeley}
}

\begin{document}
\setlength{\textfloatsep}{7pt}

\makeatletter
\let\@oldmaketitle\@maketitle%
\renewcommand{\@maketitle}{\@oldmaketitle%
    \centering
    \includegraphics[width=.95\linewidth]{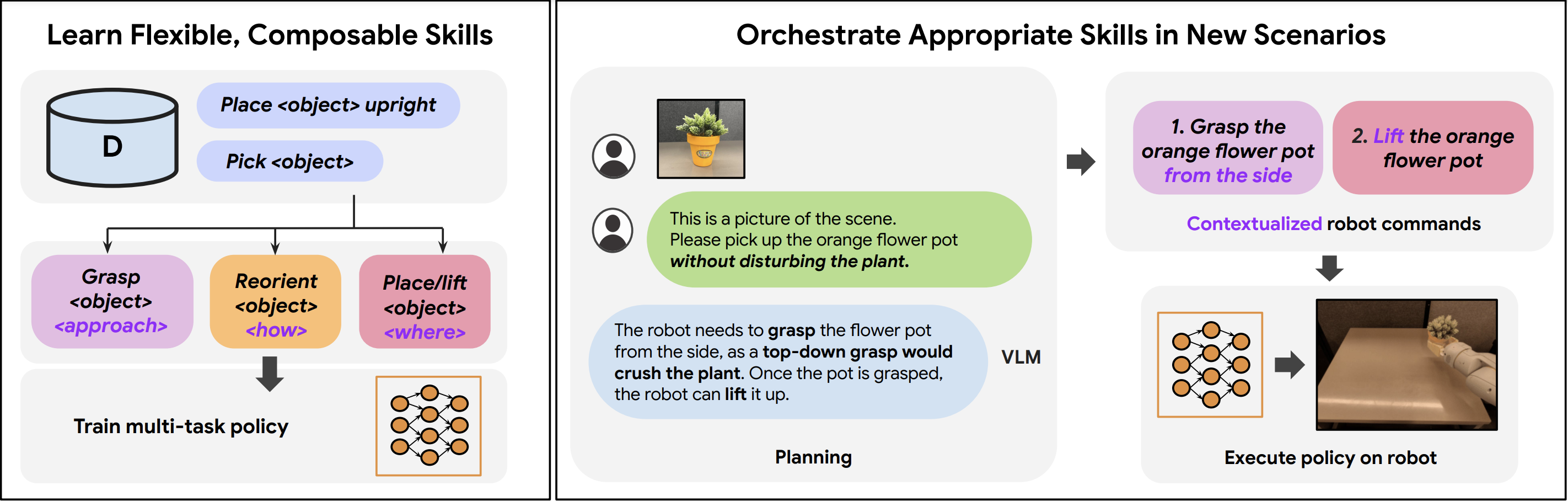}
    \captionof{figure}{\footnotesize System diagram for \metabbr. At training time, we re-annotate an offline dataset of diverse robot behaviors at training time, focusing on describing the primitive skills used to manipulate objects and, specifically, on annotating \textcolor{purple}{\textit{how}} the robot performed each skill. We then use this re-annotated dataset to train a language-conditioned low-level policy (RT-1 in our case). At inference time, when given a complex instruction like ``pick up the flower pot \emph{without disturbing the plant}'', a high-level system (VLM or human) identifies the appropriate low-level skills and determines how to perform them. This emphasis on the ``how'' enables more contextual behavior.}
    \vspace{-.4cm}
    \label{fig:teaser}
}
\makeatother

\maketitle
\thispagestyle{empty}
\pagestyle{empty}

\begin{abstract}
The complexity of the real world demands robotic systems that can intelligently adapt to unseen situations. We present \metabbr, a robot learning framework that bridges high-level, commonsense reasoning with precise, flexible low-level control. Our approach translates complex situational awareness into actionable low-level behavior through training language-grounded policies with dense annotation. By structuring policy training around fundamental, modular manipulation skills expressed in natural language, \metabbr exposes an expressive interface for humans or Vision-Language Models (VLMs) to intelligently orchestrate the robot's behavior by reasoning about the task and context. Our experiments demonstrate the skills learned via \metabbr can be combined to synthesize novel behaviors to adapt to new situations or perform completely new tasks without additional data collection or training. Project website: \videourl
\end{abstract}

\section{Introduction}
\label{sec:intro}
Consider the breadth of situations a human encounters on a daily basis, from pouring a cup of coffee in their kitchen to grabbing objects from a cluttered supply closet. Designing robot systems that can navigate these varied, nuanced scenarios is a major challenge, requiring systems that can adapt to complex and dynamic situations.
This has led many roboticists to explore learning-based solutions that may generalize better than hand-engineered ones. Imitation learning (IL) is a widely-used, data-driven approach that distills expert demonstrations into learned policies, enabling fine-grained manipulation of high-dimensional robot systems in the real world~\cite{zhao2023learning, chi2024diffusionpolicy}, and has been shown to scale with more data in language-conditioned, multi-task~\cite{jang2021bc, rt12022arxiv, RT2, driess2023palm} and even multi-robot~\cite{octo_2023, kim24openvla, doshi24crossformer} regimes.
While these works have shown remarkable promise, the resulting robot systems remain fairly limited to situations seen during training. And the span of these training scenarios is significantly narrower compared to those of other domains, such as vision and language, where large-scale supervised learning has excelled as real-world embodied data collection is significantly more expensive and bottlenecked by physical constraints.

Humans, on the other hand, can adapt to very complex situations without any previous first-hand experience. We exhibit `common sense' generalization---using our inherent understanding of complex, high-level concepts like object affordances, intuitive physics, and compositionality to adapt our past experiences intelligently as new situations arise. This deliberate, analytical thinking has been termed `System 2' processing, in contrast to reactive `System 1', reflexive low-level behaviors that are less cognitively demanding but equally essential for our behavior~\cite{kahneman2011thinking}. Emulating this blend of reasoning and reflex in robotic systems is challenging, and various approaches have been developed to bridge the gap. One notable example is SayCan~\cite{ahn2022can}, which leverages a large language model (LLM) to plan over and sequence learned policies to perform long-horizon tasks. SayCan compensates for the LLM’s lack of direct physical grounding by using the policies' value functions to assess feasibility. Moreover, this approach is limited to sequencing the original tasks demonstrated, while many realistic tasks require finer, more nuanced control of low-level policies (as illustrated in~\autoref{fig:teaser}, right side). Subsequent works have focused on enabling LLMs or VLMs to interface at a finer granularity with pre-programmed System 1 behaviors through representations like code~\cite{liang2023code} or semantic keypoints~\cite{di2024keypoint}. Another strategy that has emerged is to fine-tune VLMs on embodied data~\cite{RT2, driess2023palm}, drawing on web-scale pre-trained representations for robot control. Notably, these approaches focus on optimizing the high-level module to make the best use of a fixed set of skills, which still constrains their adaptability in unstructured, novel scenarios. Rather than modifying the System 2 reasoning layer, we pursue an orthogonal direction by focusing on making System 1 policies more flexible and responsive to high-level guidance. By designing adaptable System 1 policies, we enable seamless interaction with a fixed System 2 module—such as human or VLM-based reasoning. This adaptability allows System 1 to be dynamically adjusted in response to high-level instructions, ultimately broadening the range of tasks the system can perform and its capacity for robust, generalizable behavior.

We present \metabbr: Structured Training for EmbodiEd Reasoning, an approach for training low-level reactive policies that can be flexibly steered or directed by a higher-level reasoner, such as a human or VLM.
Our key insight for enabling this is producing \emph{dense language annotations} of the collected robot data, and training a conditional policy on granular language instructions. This policy can then be conditioned on each part of a plan generated from a high-level model (VLM/LLM), giving a combination of the respective strengths of System 1 and System 2 processes. Furthermore, this can enable adapting to new situations that require synthesizing behaviors that are not explicitly demonstrated during training. We instantiate this system using existing real-world datasets, proposing a simple automated labeling pipeline based on proprioceptive observations to extract basic object-centric manipulation skills and distill them into a low-level policy. We then propose a strategy for using a VLM to produce language directions for the low-level policy. Importantly, we show that this enables us to repurpose skills in the robot dataset in a meaningful manner at test time to handle a new situation autonomously. In summary, the contributions of this work are as follows:
\begin{itemize}
    \item We introduce \metabbr, a method that augments robot demonstration datasets with descriptive functional language annotations comprising of grasp-centric and rotation-based primitive components.
    \item We show that \metabbr enables training low-level robot policies which are significantly more flexible and steerable than prior imitation learning methods, enabling humans or pre-trained VLMs to direct low-level policies for generalizing to novel everyday manipulation tasks.
\end{itemize}
\section{Related Work}
\label{sec:relatedwork}

\noindent\textbf{Imitation Learning.}
Imitation learning (IL) has emerged as the most popular paradigm for training real-world manipulation policies~\cite{jang2021bc, chi2024diffusionpolicy, shafiullah-vqbet}; however, deploying these models in unstructured scenarios remains a challenge. Robot policies trained on human-collected demonstration data can have trouble adapting to ``out-of-distribution'' scenarios where demonstrations are sparse~\cite{mees2022hulc}. This is fundamentally due to the expensive-to-collect small scale robot data, in comparison to the web-scale text and image datasets for training today's foundation models~\cite{llama2, gpt4techreport, geminitechreport}. Due to practical constraints on obtaining more robot data, a large body of work has explored using text and vision foundation models to improve generalization in robotics from existing datasets. This includes expanding IL policies to open-world object grasping by using open-vocabulary object detection~\cite{moo} and relabeling episode-level instructions via CLIP~\cite{xiao2022dial} or other foundation models~\cite{myers2024palo,zhang2024sprint}.
Our work is closely related to aforementioned works in dataset relabeling~\cite{xiao2022dial,zhang2024sprint,myers2024palo}: our framework expands robot capabilities by relabeling different behavior modes in \emph{existing} heterogenous robot demo datasets to train more effective policies.

\noindent\textbf{Policy Conditioning for Generalization.}
Prior works have sought to enable test-time generalization by exploring expressive modalities for policy conditioning, such as goal target poses~\cite{lynch2019relay}, goal images~\cite{susie, Fang2022PlanningTP}, trajectories~\cite{gu2023rttrajectory,wen2023anypoint}, code~\cite{liang2023code}, or combinations of vision and language~\cite{vima,nasiriany2024pivot}.
However, while these modalities all show promise in action generalization, it is challenging for off-the-shelf high-level reasoning systems like VLMs or humans to plan over these creative modalities; natural language remains the main modality utilized in complex planning by state-of-the-art LLMs and VLMs.
Thus, \metabbr focuses on improving language-conditioned action prediction, by studying \emph{granular instructions} such as ``grasp from the side,'' or ``lift up,'' that can be easily composed at test-time.

\noindent\textbf{Skill Learning.} There is extensive research in utilizing learned `skills' to accelerate learning new tasks by exploring with temporally-extended, semantically meaningful action sequences~\cite{lee2019composing, lee2019learning, shiarlis2018taco, gupta2019relay, shankar2019discovering, pertsch2020spirl, singh2020parrot, sharma2020learning, dalal2021raps, nasiriany2022maple,zhang2024extract}. These works often use a hierarchical approach, where a high-level policy is learned through interaction with reinforcement learning through environment interaction to compose the learned skills~\cite{bacon2017option,nachum2018data,peng2019mcp,pertsch2020spirl, dalal2021raps, nasiriany2022maple, zhang2024extract}. EXTRACT~\cite{zhang2024extract} in particular uses VLMs to label skills from offline data to enable learning new tasks. Our work also exploits the intuition about skills being transferable to synthesize new behaviors. However, we use common-sense reasoning in off-the-shelf VLMs to choose appropriate skills for the situation without training a separate policy.

\noindent\textbf{Affordances.} Our work leverages VLMs' common-sense reasoning capabilities to plan for longer-horizon tasks and provide strategies for the robot to approach novel configurations of objects. The model does this by reasoning about how humans would approach tasks from visual image input. Prior work investigates how to represent human priors for how to act in scenes using affordances \cite{hotspots, hoi} in image space, and even shown how to deploy these on real robot systems for guiding exploration \cite{vrb, swim}. While effective, these affordances are often represented in the form of keypoint coordinates in pixel space, and make particular assumptions on the kinds of tasks to be performed. Our approach can be thought of reasoning about such affordances in language, which makes it amenable for off-the-shelf VLMs or humans to naturally interact with and opens the door to express more sophisticated descriptions than in visual space.

\section{System Design}
\label{sec:system}
We present \metabbr, a robot learning framework that aims to expand the capabilities of a robot trained on a set of expert demonstrations by extracting flexible skills from existing datasets, then relying on a module with strong reasoning capabilities to orchestrate the skills intelligently. Our system consists of two main components: low-level `System 1' skill-training and high-level `System 2' high-level reasoning. In this section, we describe design decisions for acquiring and then integrating them into a practical end-to-end system. 

\subsection{Learning Flexible, Composable Manipulation Skills}
\label{sec:chopping}
Our goal is to extract skills that can be easily reasoned about and composed by either humans of foundation models. Thus, we aim for skills that are language-indexed and object-centric, allowing a foundation model with knowledge about how the state of an object should evolve to accomplish a certain task to be able to steer despite not having direct motor control capabilities. To achieve this, we densely annotate existing datasets, then train a language-conditioned RT-1~\cite{rt12022arxiv} policy with these segmented and relabeled instructions.

Our key idea is to break down basic composable skills into \emph{semantically identifiable} categories that can be associated with a language description. As intuition, many have observed that human-collected behavior data is challenging to learn from in part due to different data collectors having different `styles' or strategies~\cite{belkhale2024data}. For example, human driving styles---aggressive, cautious, smooth, or jerky---are highly variable. Prior works have dealt with this heterogeneity by using latent variables to explain modes~\cite{morton2017simultaneouspolicylearninglatent}, new algorithms~\cite{droidIL}, or more expressive generative models~\cite{hausman2017multimodalil, chi2024diffusionpolicy}. 
We expose these different styles as adjustable parameters, allowing robots to flexibly adapt their behavior. We focus on shared, object-relational skills such as grasping, lifting, placing, and rotating. These skills, originally demonstrated using templates like \texttt{pick <object>, move <object1> near <object2>, knock <object>, place <object> upright}, can be executed with varying strategies. We identify the following key factors:

\noindent\textbf{Grasp Angle.}  Objects can be grasped in multiple stable positions, and the particular way indeed impacts the ability to perform downstream tasks. However, grasp positions are rarely prescribed (and therefore labeled), as they are often implicit. We hypothesize that controlling grasp angle can improve task composition and adaptability. We use a simple approach to label the grasp approach by manually labeling a relatively small set of `anchor' grasp poses. We then label an arbitrary grasp with the label of its nearest neighbor `anchor' pose as measured by cosine similarity. 
\begin{figure}
    \centering
    \vspace{2mm}
    \includegraphics[width=.35\textwidth]{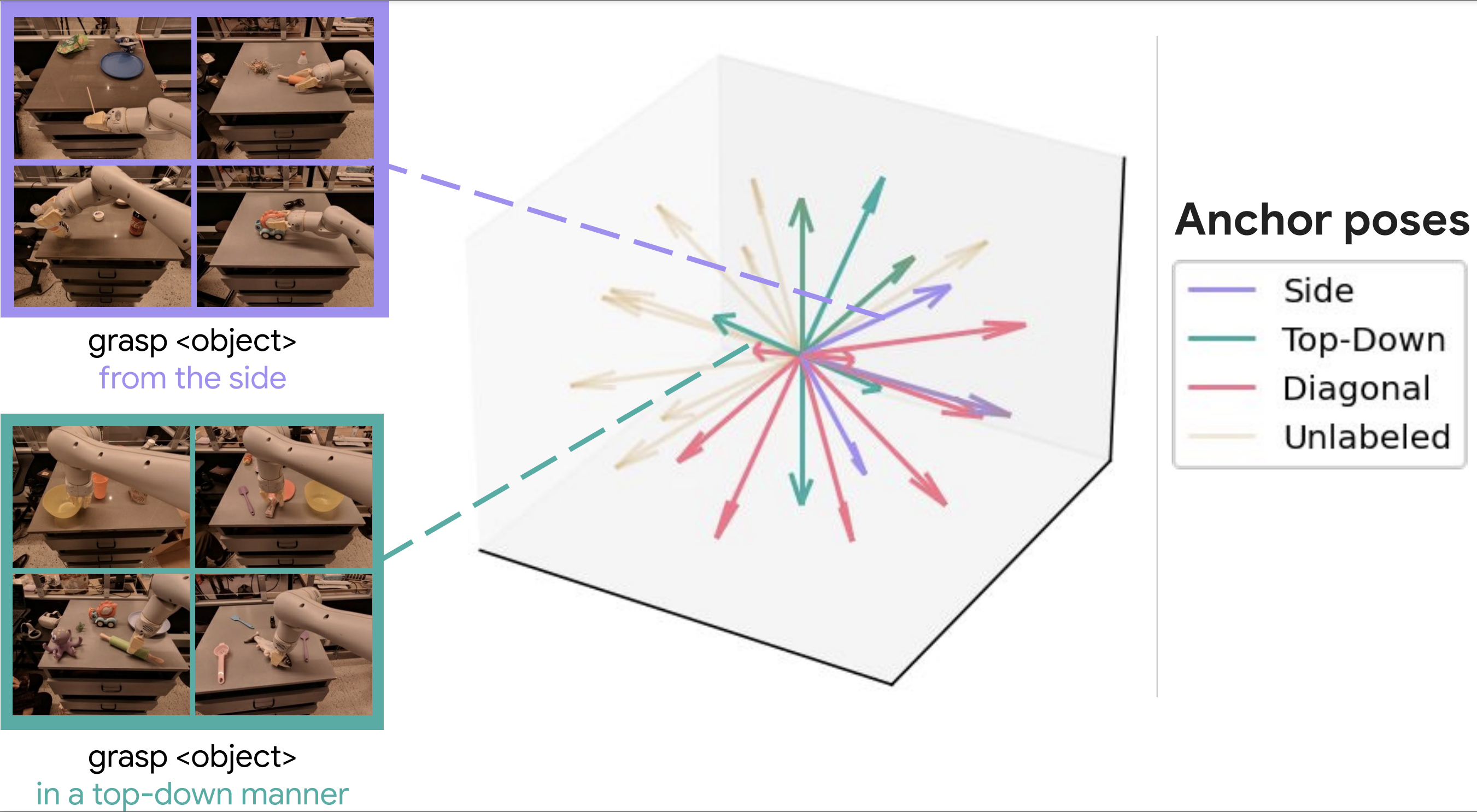}
    \caption{Anchor vectors and their semantic labels. \textcolor{purple}{Purple}, \textcolor{green}{green}, and \textcolor{pink}{pink} vectors represent side, top-down, and diagonal grasps.}
    \label{fig:graspclusters}
\end{figure}
We represent a grasp pose as a 3D unit vector by rotating $[0, 1, 0]$ by the wrist quaternion and we identify the time of a grasp where the gripper changed from fully open to fully closed. To define and label the anchor poses, we took 3D unit vectors that are linear combinations of the elementary 3D basis vectors (resulting in 27 directions). We then clustered 1000 grasps from our dataset and visualized the clusters in order to label them. This only requires visualizing and labeling roughly 20 clusters, but then can be used to automatically label the entire dataset of 70K demonstrations. In the grasp data, we identify three distinct modes via inspecting the grasp images: top-down grasps, side grasps, and diagonal grasps (visualized in~\autoref{fig:graspclusters}). The sub-trajectory starting from the beginning of the demonstration to the time of the grasp identified is relabeled to \texttt{grasp the <object> in a <grasp approach>}, where \texttt{<object>} is from the original instruction and \texttt{<grasp approach>} is from the anchor's label.

\noindent\textbf{Reorientation.} Another mode of behavior identified in the dataset is of reorienting objects. In order to identify and label these reorientations, we first label the wrist orientation for every timestep where the gripper is fully closed. Then if the gripper orientation switches between two of the modes (as labeled in \textbf{Grasp Angle}), we label the sub-trajectory preceding it as \texttt{reorient the <object> <direction>}, where \texttt{<object>} again is from the original instruction and \texttt{<direction>} indicates whether the object is rotated from upright to horizontal or vice versa. 

\noindent\textbf{Lifting/Placing.} Complementing grasping, we label whether the object was lifted or placed at the end of completing the original task. Lifting allows the robot to continue in-hand manipulation without setting the object down or provides the ability to gain additional clearance. If the object is still held at the end of the episode and the gripper moves vertically upward (common for originally-labeled \texttt{pick <object>} episodes), we label the final sub-trajectory as \texttt{hold and lift the <object>}. If not, similar to identifying grasps, we identify the time of placing using the gripper state and label this sub-trajectory as \texttt{place the <object>}. As this is separate from reorientation, placing implies setting the object down while maintaining its orientation. 

\subsection{Orchestrating Learned Skills} A key capability of the System 2 component is its ability to reason about the visual observation of the scene, the task description, and the robot's low-level skills to effectively select and sequence appropriate actions. While a human can perform this reasoning, we also demonstrate how it can be automated using a VLM. Our automated System 2 component is implemented as a code-writing VLM agent, enabling it to autonomously execute verbalized plans without additional modules or a human in the loop. To facilitate this, we define an API for the action primitives accessible by the VLM to interface with the System 1, reactive low-level RT-1 policy skills as described in \autoref{sec:chopping}. The API is based on translating the language commands into a simple API that the VLM agent can access. This breakdown is based on \emph{what} the robot should do and \emph{how} to do it. Each primitive skill (i.e. grasping, rotating, lifting, placing) is represented by a function with a keyword argument modifying \emph{how} that primitive is accomplished (i.e. \texttt{grasp(object, "top-down"}). Internally, the API translates this code into the corresponding natural language the RT-1 policy was trained on. Following Arenas et al.~\cite{arenas2023prompt}, we use a system prompt to tell the VLM to control its physical embodiment through code, then provide the robot's visual observation of the scene and a description of the high-level task. One benefit of this design is its modularity and interpretability. The prompt can be modified to tailor behavior according to the specific situation the robot is in, the generated plans can be viewed before execution.
The exact system prompt we use in all experiments and example outputs produced by the model can be found on our project website: \url{https://lauramsmith.github.io/steer}. 

\begin{figure}[t!]
\centering
\includegraphics[width=.4\textwidth]{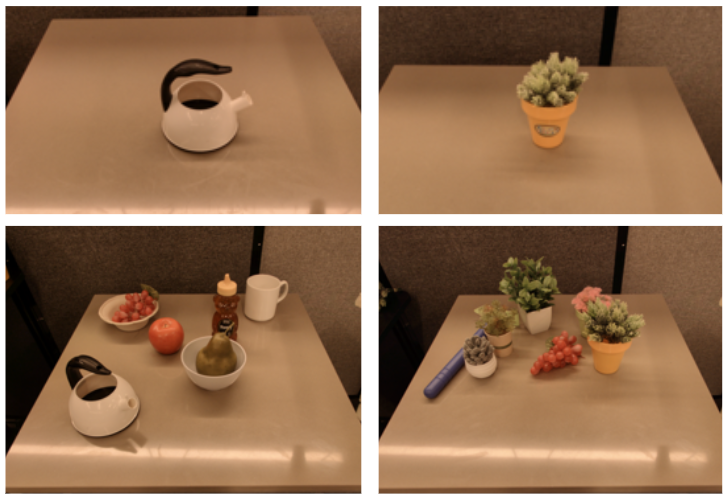}
\caption{\textbf{Sample initial conditions} for the new object-grasping scenarios evaluated. (top left) a kettle with a handle extending above it, (top right) a potted plant, (bottom) 2/15 scenes for \texttt{Fruit in Clutter}. The kettle should be grasped over top. In order to avoid disturbing the plant, the flower pot should be grasped around its body. Lastly, the fruits should be grasped while avoiding knocking over the other objects in the scene.}
\label{fig:graspinginitvis}
\end{figure}

\begin{figure}
    \centering
    \includegraphics[width=.4\textwidth]{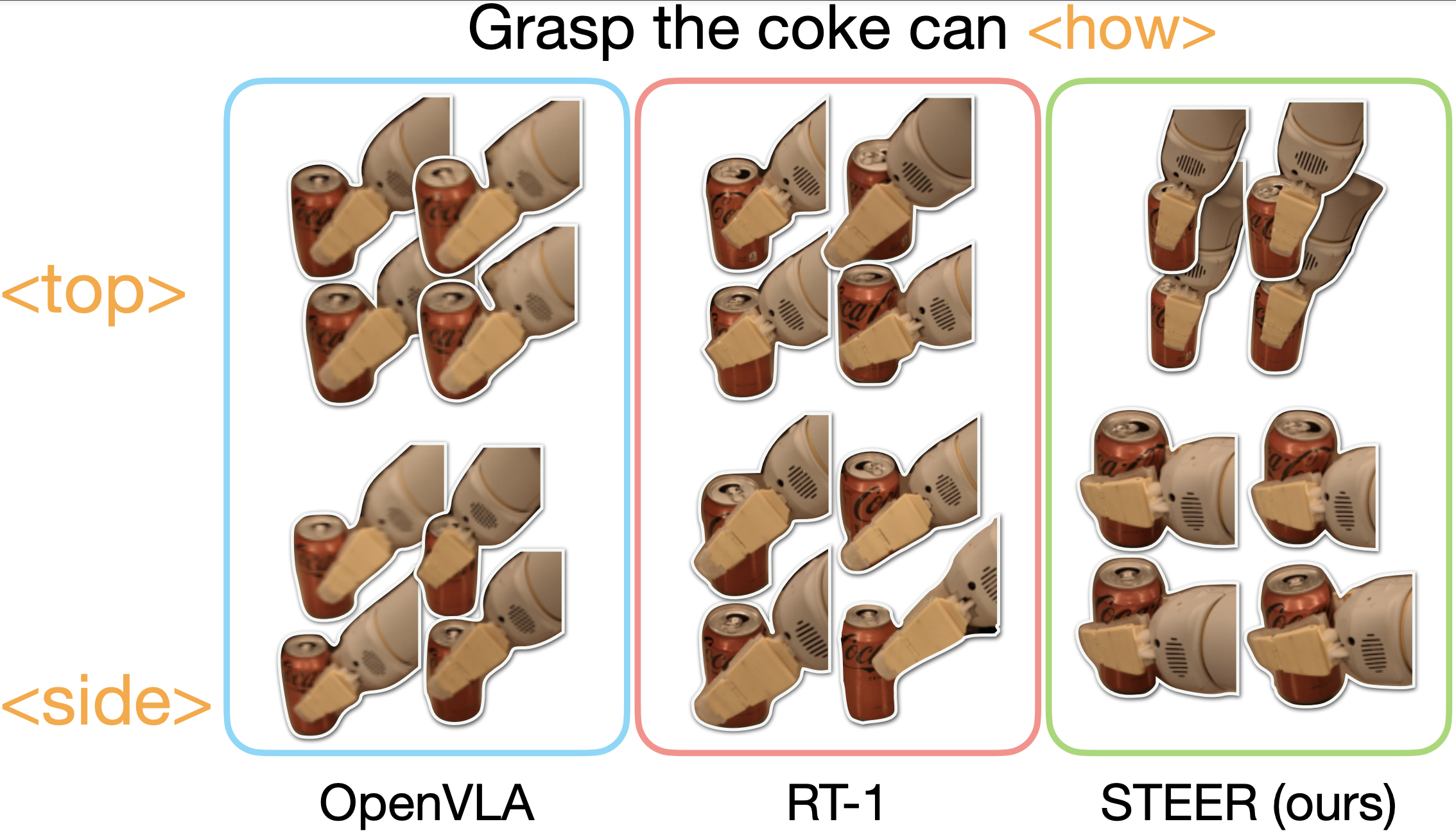}
    \caption{\textbf{Grasp steerability} of OpenVLA, RT-1, and \metabbr. We test the steerability to grasp an object in different ways that would be appropriate for different, unseen tasks, e.g., in order to pour out of the Coke can, the robot should grasp the can around its body. When prompted to “Grasp the Coke can” from the top (top row) versus the side (bottom row), models without dense annotation show no perceivable change, while our densely labeled model adjusts its behavior, enabling new downstream tasks.}
    \label{fig:graspanglevis}
\end{figure}
\begin{figure*}[ht]
    \centering
    \begin{subfigure}[t]{0.3\textwidth}
        \includegraphics[width=\linewidth]{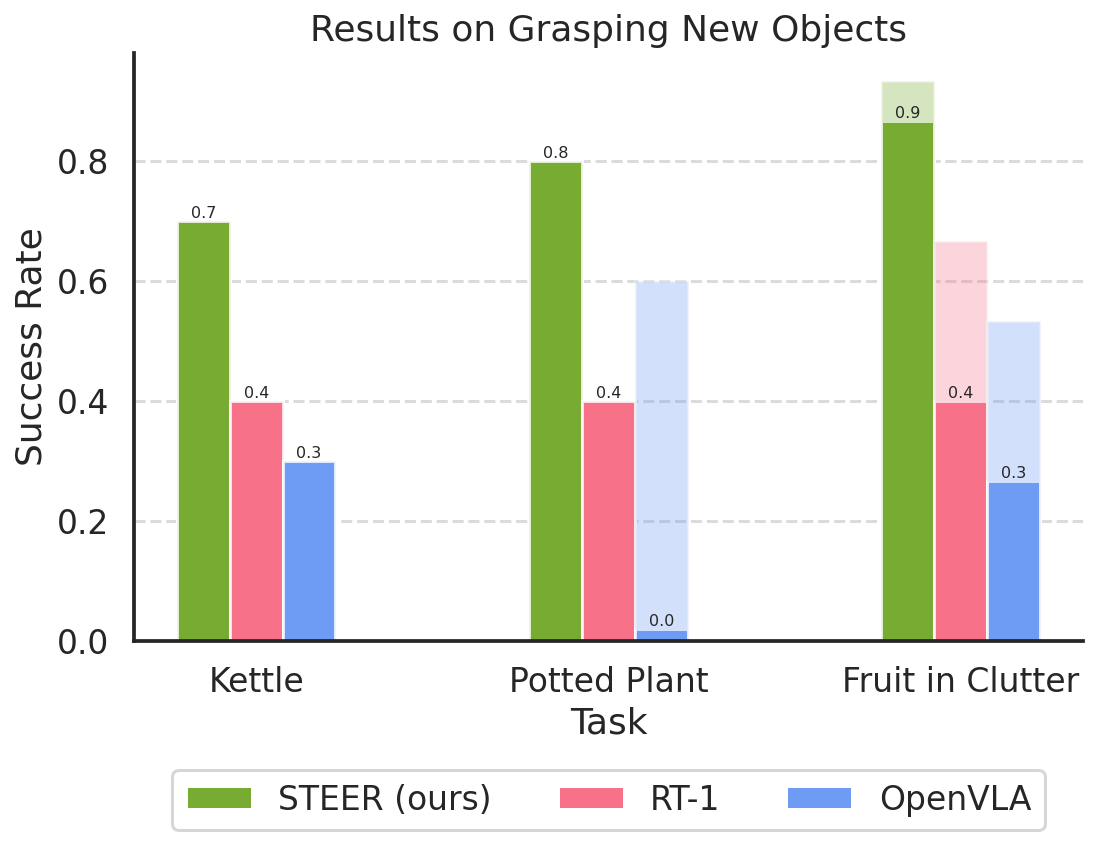}
        \caption{Grasping results. Full successes are in solid colors. Partial successes (defined as grasping the object with undesired side effects like disturbing the plant or other objects) are in light colors. For each method, we do 20 trials for \texttt{Kettle}, 10 trials for \texttt{Potted Plant}, and 15 trials/configurations of \texttt{Fruit in Clutter}.}
        \label{fig:grasping_plot}
    \end{subfigure}
    \begin{subfigure}[t]{0.3\textwidth}
        \centering
        \includegraphics[width=\linewidth]{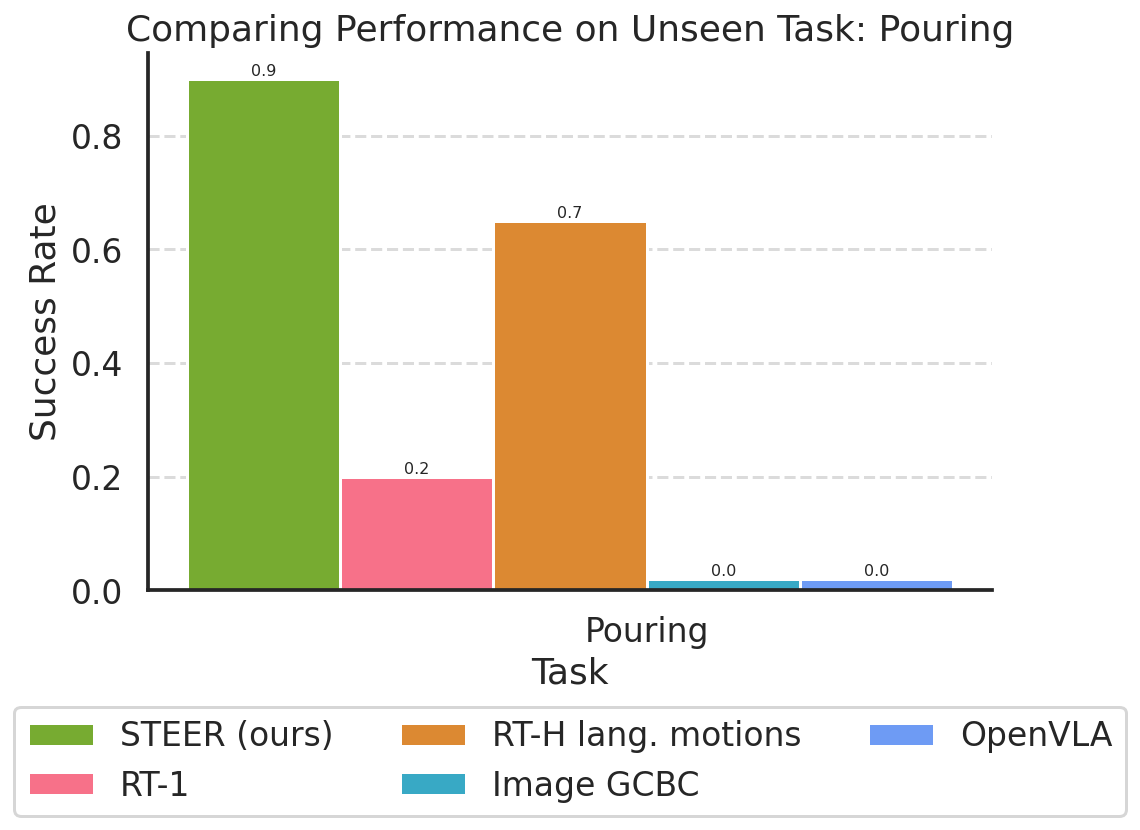}
        \caption{New task results. We run each method 10 times, comparing the upper bound on performance of the low-level capabilities afforded by each model to perform the task. Therefore, we use closed-loop human guidance for language-instructed models and granular goal images for the image-conditioned policy.}
        \label{fig:new_task_plot}
    \end{subfigure}
    \begin{subfigure}[t]{0.3\textwidth}
        \centering
        \includegraphics[width=\linewidth]{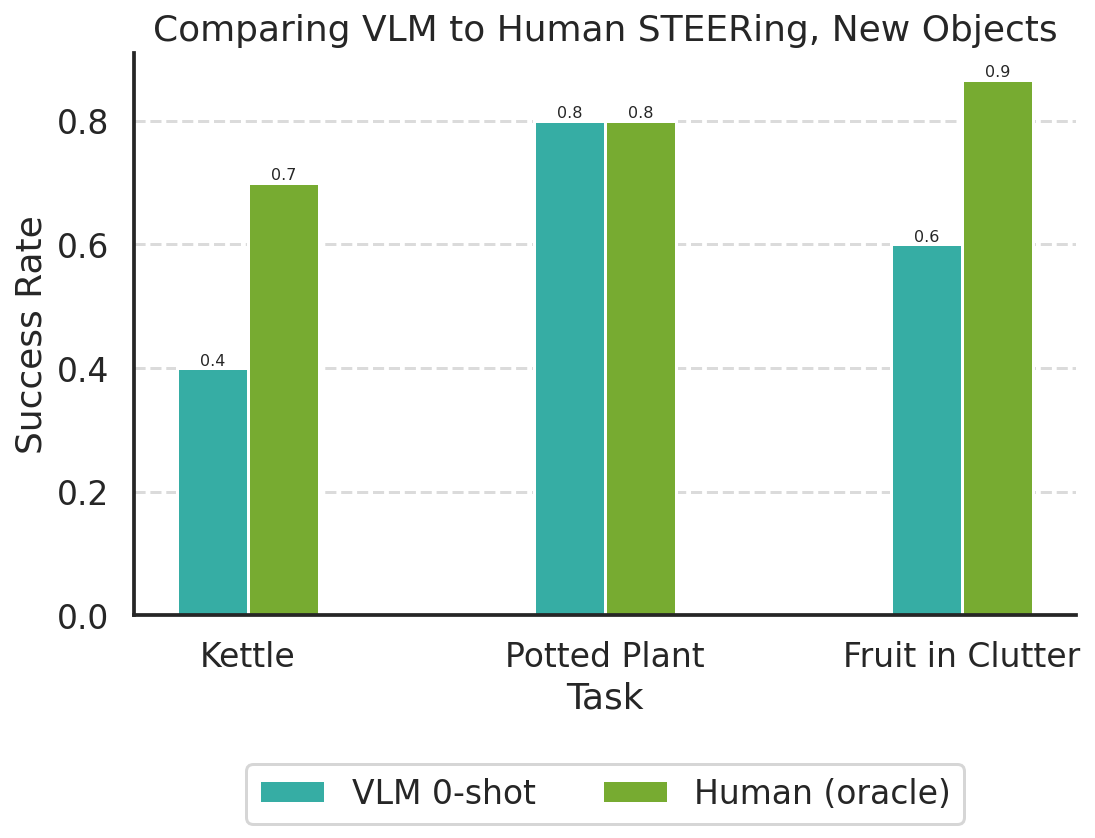}
        \caption{We report the success rate of a VLM controlling the robot compared to the human operator and find that the VLM produces reasonable plans but there is low-level policy failure due to sensitivity of the low-level policy to language inputs.}
        \label{fig:gemini_0shot}
    \end{subfigure}
    \caption{Results on grasping in unseen scenarios and performing a new task, with human or VLM guidance. We find that by having access to and being able to reason about extracted low-level strategies enables higher success in OOD scenarios than the baseline RT-1 model and a state-of-the-art VLA.}
    \vspace{-7mm}
\end{figure*}

\section{Experiments}
\label{sec:experiments}
To understand the efficacy afforded by multiple strategies extracted and learned from the offline data, we test whether our model enables more effective grasping in unseen scenarios. We test this by manipulating objects that do not appear in the offline dataset \emph{and} require specific grasp strategies to succeed. We then test \metabbr's ability to perform new behaviors---requiring complex reasoning \textit{and} reliable motor control. This is first demonstrated by having a human expert create a plan for \metabbr, to show our design decisions lead to composable primitives. We then leverage a VLM to automate the planning. We aim to answer the following concrete questions through our experiments:
\begin{enumerate}
\item Does learning multiple modes of behaviors used to solve a task improve the adaptability of a robotic manipulation system to novel situations?
\item Can combining extracted skills from heterogeneous human demos enable entirely new tasks?
\item To what degree can a state-of-the-art VLM plan orchestrate these skills autonomously?
\end{enumerate}

\subsection{Experimental Setup}
We use a mobile manipulator with a 7 DoF arm, a two-fingered gripper, and a mobile base, as used in \mbox{RT-1}~\cite{rt12022arxiv}. We target a tabletop manipulation environment, where objects on a counter need to be moved or arranged according to natural language instruction. 
We use the multi-task (6 semantic categories) demonstration dataset used in RT-1~\cite{rt12022arxiv} (70K demonstrations) and the dataset of grasping-only data featured in MOO~\cite{moo} (15K demonstrations). As we discuss in \autoref{sec:system}, the exact architectures of the low-level (System 1) and high-level (System 2) components can vary, as long as the high-level component can reason over skills expressed in language. We choose RT-1~\cite{rt12022arxiv} for our System 1 component and Gemini 1.5 Pro~\cite{gemini15} in VLM experiments. Videos and setup details are available on the project website: \url{https://lauramsmith.github.io/steer}.

\subsection{Improving Test-Time Adaptability} 

\noindent\textbf{Setup.} One of our central hypotheses is that explicitly extracting, labeling, and training expressive primitives from heterogeneous demonstrations affords our system with improved robustness. When faced with unfamiliar situations at test time, we expect some skills to generalize better or be more suitable than others. To test this, we present the robot with three unseen object-grasping scenarios (sample initial conditions shown in~\autoref{fig:graspinginitvis}): a kettle with a handle extending above it, a potted plant, and grasping a fruit out of clutter.
We choose unseen test objects specifically to simulate challenging real-world settings in which na\"ive grasping without a strategy is unlikely to succeed. 

We first compare \metabbr against the baseline RT-1~\cite{rt12022arxiv} model, with the same architecture, trained on the same demo data as \metabbr, but with the \emph{original language instructions}. We condition \metabbr on the templated language it is trained on, with the grasp strategy chosen by a human in these experiments. We condition RT-1 on the templated language it is trained on, i.e. ``pick <object>''. For \metabbr and RT-1, we train with a 50/50 split of the RT-1 and MOO datasets since MOO comprises diverse grasping-only data. We then compare to OpenVLA~\cite{kim24openvla}, a model that fine-tunes a VLM~\cite{llama2} on robot data from the Open X-Embodiment dataset~\cite{open_x_embodiment_rt_x_2023}. The OXE dataset includes the same demo data that RT-1 and \metabbr were trained on, in addition to data from various other robots and tasks. This is meant to test whether fine-tuning a VLM on robot data is sufficient to elicit the desired reasoning and downstream execution capabilities for these tasks as opposed to explicit re-annotation and direction by a high-level module. We condition OpenVLA on the same language, i.e., ``pick orange flower pot without disturbing the plant" or ``pick apple while avoiding the other objects" that we also provide our VLM to sufficiently define the task in~\autoref{sec:vlmexps}.

\noindent\textbf{Results.} We report the success rates in \autoref{fig:grasping_plot}. RT-1 occasionally succeeds, but exhibits different strategies and we observe that failures are often caused by a sub-optimal approach. For example, when approaching the potted plant without a direct side grasp, the pot is prone to falling or grasping the plant leaves. OpenVLA performed similarly to RT-1, demonstrating that additional web data does not necessarily lead to sufficiently strong embodied reasoning about \emph{how} to grasp in a new scenario where a particular approach is evidently necessary. For example, we find that OpenVLA often picks the potted plant up, but does not respect the language instruction of picking up the flower pot \emph{without disturbing the plant} and grasps from above around the plant leaves. We further test whether it is \emph{possible} with human guidance to steer RT-1 and OpenVLA to grasp differently through the same language prompting as \metabbr. As shown in~\autoref{fig:graspanglevis}, \metabbr clearly changes its grasp strategy based on language prompting, whereas without re-annotation, neither OpenVLA nor RT-1 adjusts its behavior with more descriptive conditioning. This highlights that decomposing the grasp strategies and exploiting the most suitable one as we do in \metabbr is necessary in this case to coax a model to generalize correctly.

\subsection{Performing Novel Behaviors} 
\noindent\textbf{Setup.} Having demonstrated that \metabbr learns more flexible skill primitives, we study how well we can engineer behavior for a new everyday task without collecting new demonstrations or additional fine-tuning. To test this, we consider an everyday task, pouring, that is out of the distribution of demonstrated tasks but should be achievable with the motions that exist in the data. Pouring requires grasping the cup from the side, such that the robot can easily tilt the cup once lifted---avoiding singularities or spilling onto the robot---then setting the cup back down onto the table upright.

We perform an extensive evaluation on the pouring task, comparing against the \textbf{best-case} version of each baseline and comparison. Each method exposes a different control interface---for example, RT-1 is commanded with the natural language instructions it has been trained on, while a BC policy conditioned on goal images is commanded by giving images of the sub-goals required to reach the desired end state. For each method, if human assistance is used, the human coaxes the model to perform the task by providing what they deem to be the best command that method supports, in a closed-loop fashion. The human is not allowed to move the objects or robot on their own, and a trial is halted if the robot reaches an irrecoverable state (e.g. objects fall off the table). For this task, we train RT-1 and \metabbr with a mixing ratio proportional to the respective dataset sizes (roughly 85\% RT-1 dataset 15\% MOO dataset) as we are not testing generalization to new \emph{objects}, rather maneuvering seen objects in new ways. We compare to the following 4 baselines and prior methods:
\begin{enumerate}
    \item RT-1~\cite{rt12022arxiv}, which acts as our baseline set of action primitives given by the original tasks in the datasets.
    \item Language motions from RT-H~\cite{belkhale2024rt}, defined by narrating end-effector movement to give language like \texttt{move arm left} and \texttt{rotate arm right}.
    \item A goal-image conditioned variant of RT-1. This tests whether language is a better abstraction than goal images. For this comparison, we first perform a demonstration of the new task, then run a goal-image conditioned policy passing images from that demonstration as subgoals at the granularity of our extracted skills.
    \item OpenVLA~\cite{kim24openvla}, to compare to a state-of-the-art model pre-trained on web data and fine-tuned on robot data.
\end{enumerate}

\noindent\textbf{Results.} Human orchestration with a \metabbr policy achieves a 90\% success rate on pouring as compared to 70\% with a policy trained with language motions from RT-H (\autoref{fig:new_task_plot}). In comparison, baseline RT-1 cannot complete the task because it is not trained to reorient objects. Instead, it is trained to "knock" over objects and place objects upright. If the policy successfully places the cup back upright after knocking, though, we consider the trial a half-success since knocking the cup onto the table is not a desirable pouring motion. The goal image baseline, despite having demonstration subgoal images from the same starting positions, fails to perform the task successfully. We observe that the policy is brittle and appears to match the arm pose in the subgoals, rather than manipulate the object state as prescribed by the goal image. OpenVLA, despite having access to the same underlying robot demonstration data and VLM pre-training, does not generalize to the new motion by stitching together the appropriate motions. Instead, we observe that it attempts to pick up and often knock over the cup. Despite the very dense language motion instructions proposed by RT-H achieving a high success rate, orchestration is significantly more cumbersome as it requires tight closed-loop guidance. This is evidenced by requiring on average 15 instructions of the type "move arm forward and left" done in-the-loop as opposed to 5 simple commands that can be executed open-loop by \metabbr (i.e., "grasp pink cup from the side", "lift pink cup", "reorient the pink cup to be horizontal", "reorient the pink cup to be vertical", "place the pink cup").

\subsection{VLM Orchestration}
\label{sec:vlmexps}
Now, we test whether a \emph{VLM} can effectively select or sequence appropriate skills afforded by \metabbr by reasoning about the context, in the visual observation and task description, as well as the skills exposed through the API \emph{without any examples} (i.e. 0-shot). For these experiments, we compare to human orchestration of the same \metabbr policy as an upper bound on the performance. For each trial, we query the VLM with the initial scene, task description, and maintain the same system prompt. Exact inputs and outputs for all experiments are on the project website.

\noindent\textbf{Seen task, new scenarios.} We see that the VLM successfully produces the same high-level plans as the human expert very reliably for the grasping tasks. However, as shown in~\autoref{fig:gemini_0shot} we see that there is a degradation in end-to-end task performance compared to human orchestration when executing the code produced by the VLM, and we analyze these failures.
For the kettle picking task, we note that the low-level policy appears to be sensitive to the specific naming of objects. That is, the VLM often produced code to grasp the `black and white kettle` from the top instead of grasping the `black and white object` from the top, and with further analysis find that this instruction has a noticeable degradation across all low-level language-conditioned policies. So, while the VLM reasonably commands the policy to grasp from above, the low-level policy is less reliable. We expect this to be improved with denser annotation or augmentation on the entity-level, whereas \metabbr is concerned with the motion-level. For the \texttt{Fruit in Clutter} grasping task, the VLM did not always command the appropriate action and we suspect that similar object naming references (`red apple' instead of `apple') impact the low-level policy performance.  

\noindent\textbf{Seen objects, new task.}  We observe that without any examples, the VLM correctly identifies the grasp location to pour from the cup,in a manner that a human would perform the task. It then recognizes that it must reorient it, then reorient it back in order to place it back upright on the table. A common failure mode is that the VLM misunderstands that 90 degree rotation does not invert the cup. However, this artifact luckily does not affect performance on \emph{this} task. The VLM succeeded in $6/10$ trials for zero-shot pouring.

\subsection{Additional Experiments}
\begin{figure}
    \vspace{2mm}
    \centering
    \includegraphics[width=\linewidth]{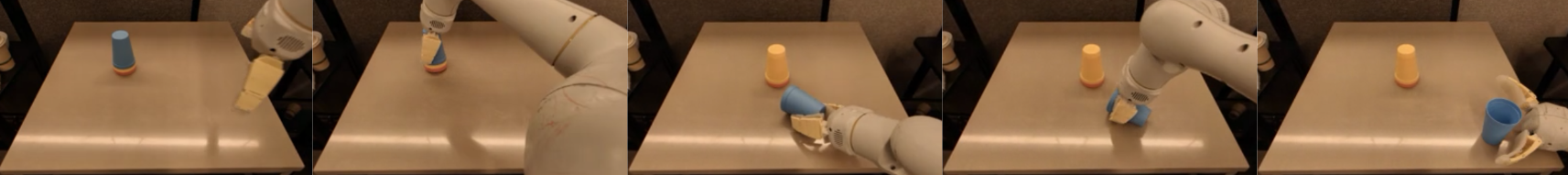}
    \caption{Rollout of \metabbr solving multi-step manipulation tasks with human guidance and steerable manipulation.}
    \label{fig:flipcup}
\end{figure}
\noindent\textbf{Self-improvement with in-context learning.} For the pouring task, the VLM orchestrated policy succeeded in $6/10$ trials. We tested whether feeding the VLM-generated programs that resulted in on-robot success as examples in the prompt would lead to a higher success rate. Indeed, with the self-generated in-context examples, the newly generated programs succeeded in $8/10$ trials for the same pouring task, improving upon the $6/10$ in 0-shot. This only requires human labels for task success and no model fine-tuning.

\noindent\textbf{Cup unstacking and flipping.} We also test our ability to use \metabbr to perform a longer-horizon task of unstacking and reorienting a cup upright, so as to be able to dispense a drink. This task also requires grasping \textit{with intention} for the future object reorientation steps. We demonstrate that we can guide \metabbr to perform this new task (see~\autoref{fig:flipcup}).

\section{Discussion}
\label{sec:conclusion}

We presented \metabbr, a robotic system that can follow natural language instructions for manipulation. Rather than collecting data for new tasks, \metabbr uses a novel methodology for relabeling \emph{existing} data with flexible and composable manipulation primitives. This relabeled data is used to train a small language-conditioned policy, which can be controlled using either a high-level VLM agent or a human. \metabbr can be \emph{steered} to perform specific manipulation tasks, which in conjunction with the commonsense-reasoning afforded by a VLM agent can perform intelligent multi-step manipulation without ever collecting new data. We report that our simple recipe can outperform large, end-to-end models like OpenVLA (despite using a $100\times$ smaller model and data). Since the performance of \metabbr is driven by our re-annotation of behavior modes, this suggests existing robot datasets could also be re-annotated to produce more steerable action primitives, with little changes to the model architecture and training pipelines. In the future, it would be useful to scale up the discovery and labeling of dataset attributes, investigate automatic relabeling, and more directly optimize annotations to maximize the high-level agent's skill composability.

\section{Acknowledgements}
Laura Smith is supported by the Google PhD Fellowship. We are grateful to Peng Xu, Alex Kim, Paul Wohlhart, and Justice Carbajal for help with robot and compute infrastructures. We thank Jonathan Tompson, Annie Xie, Sudeep Dasari, Chen Wang, Jason Ma, Danny Driess, Soroush Nasiriany, Dorsa Sadigh, Philip Ball, Kelvin Xu, and Joey Hejna for valuable discussions. Finally, we thank Jacky Liang and Carolina Parada for feedback on the paper. 

\balance
\bibliography{references}

\begin{thebibliography}{10}
\providecommand{\url}[1]{#1}
\csname url@samestyle\endcsname
\providecommand{\newblock}{\relax}
\providecommand{\bibinfo}[2]{#2}
\providecommand{\BIBentrySTDinterwordspacing}{\spaceskip=0pt\relax}
\providecommand{\BIBentryALTinterwordstretchfactor}{4}
\providecommand{\BIBentryALTinterwordspacing}{\spaceskip=\fontdimen2\font plus
\BIBentryALTinterwordstretchfactor\fontdimen3\font minus \fontdimen4\font\relax}
\providecommand{\BIBforeignlanguage}[2]{{%
\expandafter\ifx\csname l@#1\endcsname\relax
\typeout{** WARNING: IEEEtran.bst: No hyphenation pattern has been}%
\typeout{** loaded for the language `#1'. Using the pattern for}%
\typeout{** the default language instead.}%
\else
\language=\csname l@#1\endcsname
\fi
#2}}
\providecommand{\BIBdecl}{\relax}
\BIBdecl

\bibitem{zhao2023learning}
T.~Z. Zhao, V.~Kumar, S.~Levine, and C.~Finn, ``Learning fine-grained bimanual manipulation with low-cost hardware,'' \emph{Robotics: Science and Systems (RSS)}, 2023.

\bibitem{chi2024diffusionpolicy}
C.~Chi, Z.~Xu, S.~Feng, E.~Cousineau, Y.~Du, B.~Burchfiel, R.~Tedrake, and S.~Song, ``Diffusion policy: Visuomotor policy learning via action diffusion,'' \emph{The International Journal of Robotics Research}, 2024.

\bibitem{jang2021bc}
\BIBentryALTinterwordspacing
E.~Jang, A.~Irpan, M.~Khansari, D.~Kappler, F.~Ebert, C.~Lynch, S.~Levine, and C.~Finn, ``{BC}-z: Zero-shot task generalization with robotic imitation learning,'' in \emph{5th Annual Conference on Robot Learning}, 2021. [Online]. Available: \url{https://openreview.net/forum?id=8kbp23tSGYv}
\BIBentrySTDinterwordspacing

\bibitem{rt12022arxiv}
A.~Brohan, N.~Brown, J.~Carbajal, Y.~Chebotar, J.~Dabis, C.~Finn, K.~Gopalakrishnan, K.~Hausman, A.~Herzog, J.~Hsu, J.~Ibarz, B.~Ichter, A.~Irpan, T.~Jackson, S.~Jesmonth, N.~Joshi, R.~Julian, D.~Kalashnikov, Y.~Kuang, I.~Leal, K.-H. Lee, S.~Levine, Y.~Lu, U.~Malla, D.~Manjunath, I.~Mordatch, O.~Nachum, C.~Parada, J.~Peralta, E.~Perez, K.~Pertsch, J.~Quiambao, K.~Rao, M.~Ryoo, G.~Salazar, P.~Sanketi, K.~Sayed, J.~Singh, S.~Sontakke, A.~Stone, C.~Tan, H.~Tran, V.~Vanhoucke, S.~Vega, Q.~Vuong, F.~Xia, T.~Xiao, P.~Xu, S.~Xu, T.~Yu, and B.~Zitkovich, ``Rt-1: Robotics transformer for real-world control at scale,'' in \emph{arXiv preprint arXiv:2212.06817}, 2022.

\bibitem{RT2}
A.~Brohan, N.~Brown, J.~Carbajal, Y.~Chebotar, X.~Chen, K.~Choromanski, T.~Ding, D.~Driess, A.~Dubey, C.~Finn \emph{et~al.}, ``Rt-2: Vision-language-action models transfer web knowledge to robotic control,'' \emph{arXiv preprint arXiv:2307.15818}, 2023.

\bibitem{driess2023palm}
D.~Driess, F.~Xia, M.~S. Sajjadi, C.~Lynch, A.~Chowdhery, B.~Ichter, A.~Wahid, J.~Tompson, Q.~Vuong, T.~Yu \emph{et~al.}, ``Palm-e: An embodied multimodal language model,'' \emph{arXiv preprint arXiv:2303.03378}, 2023.

\bibitem{octo_2023}
{Octo Model Team}, D.~Ghosh, H.~Walke, K.~Pertsch, K.~Black, O.~Mees, S.~Dasari, J.~Hejna, C.~Xu, J.~Luo, T.~Kreiman, Y.~Tan, L.~Y. Chen, P.~Sanketi, Q.~Vuong, T.~Xiao, D.~Sadigh, C.~Finn, and S.~Levine, ``Octo: An open-source generalist robot policy,'' in \emph{Proceedings of Robotics: Science and Systems}, Delft, Netherlands, 2024.

\bibitem{kim24openvla}
M.~Kim, K.~Pertsch, S.~Karamcheti, T.~Xiao, A.~Balakrishna, S.~Nair, R.~Rafailov, E.~Foster, G.~Lam, P.~Sanketi, Q.~Vuong, T.~Kollar, B.~Burchfiel, R.~Tedrake, D.~Sadigh, S.~Levine, P.~Liang, and C.~Finn, ``Openvla: An open-source vision-language-action model,'' \emph{arXiv preprint arXiv:2406.09246}, 2024.

\bibitem{doshi24crossformer}
R.~Doshi, H.~Walke, O.~Mees, S.~Dasari, and S.~Levine, ``Scaling cross-embodied learning: One policy for manipulation, navigation, locomotion and aviation,'' \emph{arXiv preprint arXiv:2408.11812}, 2024.

\bibitem{kahneman2011thinking}
\BIBentryALTinterwordspacing
D.~Kahneman, \emph{Thinking, fast and slow}.\hskip 1em plus 0.5em minus 0.4em\relax New York: Farrar, Straus and Giroux, 2011. [Online]. Available: \url{https://www.amazon.de/Thinking-Fast-Slow-Daniel-Kahneman/dp/0374275637/ref=wl_it_dp_o_pdT1_nS_nC?ie=UTF8&colid=151193SNGKJT9&coliid=I3OCESLZCVDFL7}
\BIBentrySTDinterwordspacing

\bibitem{ahn2022can}
M.~Ahn, A.~Brohan, N.~Brown, Y.~Chebotar, O.~Cortes, B.~David, C.~Finn, C.~Fu, K.~Gopalakrishnan, K.~Hausman \emph{et~al.}, ``Do as i can, not as i say: Grounding language in robotic affordances,'' in \emph{6th Annual Conference on Robot Learning}, 2022.

\bibitem{liang2023code}
J.~Liang, W.~Huang, F.~Xia, P.~Xu, K.~Hausman, B.~Ichter, P.~Florence, and A.~Zeng, ``Code as policies: Language model programs for embodied control,'' in \emph{2023 IEEE International Conference on Robotics and Automation (ICRA)}.\hskip 1em plus 0.5em minus 0.4em\relax IEEE, 2023, pp. 9493--9500.

\bibitem{di2024keypoint}
N.~Di~Palo and E.~Johns, ``Keypoint action tokens enable in-context imitation learning in robotics,'' \emph{arXiv preprint arXiv:2403.19578}, 2024.

\bibitem{shafiullah-vqbet}
\BIBentryALTinterwordspacing
S.~Lee, Y.~Wang, H.~Etukuru, H.~J. Kim, N.~M.~M. Shafiullah, and L.~Pinto, ``Behavior generation with latent actions,'' 2024. [Online]. Available: \url{https://arxiv.org/abs/2403.03181}
\BIBentrySTDinterwordspacing

\bibitem{mees2022hulc}
O.~Mees, L.~Hermann, and W.~Burgard, ``What matters in language conditioned robotic imitation learning over unstructured data,'' \emph{IEEE Robotics and Automation Letters (RA-L)}, vol.~7, no.~4, pp. 11\,205--11\,212, 2022.

\bibitem{llama2}
\BIBentryALTinterwordspacing
H.~Touvron, L.~Martin, K.~Stone, P.~Albert, A.~Almahairi, Y.~Babaei, N.~Bashlykov, S.~Batra, P.~Bhargava, S.~Bhosale, D.~Bikel, L.~Blecher, C.~C. Ferrer, M.~Chen, G.~Cucurull, D.~Esiobu, J.~Fernandes, J.~Fu, W.~Fu, B.~Fuller, C.~Gao, V.~Goswami, N.~Goyal, A.~Hartshorn, S.~Hosseini, R.~Hou, H.~Inan, M.~Kardas, V.~Kerkez, M.~Khabsa, I.~Kloumann, A.~Korenev, P.~S. Koura, M.-A. Lachaux, T.~Lavril, J.~Lee, D.~Liskovich, Y.~Lu, Y.~Mao, X.~Martinet, T.~Mihaylov, P.~Mishra, I.~Molybog, Y.~Nie, A.~Poulton, J.~Reizenstein, R.~Rungta, K.~Saladi, A.~Schelten, R.~Silva, E.~M. Smith, R.~Subramanian, X.~E. Tan, B.~Tang, R.~Taylor, A.~Williams, J.~X. Kuan, P.~Xu, Z.~Yan, I.~Zarov, Y.~Zhang, A.~Fan, M.~Kambadur, S.~Narang, A.~Rodriguez, R.~Stojnic, S.~Edunov, and T.~Scialom, ``Llama 2: Open foundation and fine-tuned chat models,'' 2023. [Online]. Available: \url{https://arxiv.org/abs/2307.09288}
\BIBentrySTDinterwordspacing

\bibitem{gpt4techreport}
\BIBentryALTinterwordspacing
OpenAI \emph{et~al.}, ``Gpt-4 technical report,'' 2024. [Online]. Available: \url{https://arxiv.org/abs/2303.08774}
\BIBentrySTDinterwordspacing

\bibitem{geminitechreport}
\BIBentryALTinterwordspacing
G.~Team \emph{et~al.}, ``Gemini: A family of highly capable multimodal models,'' 2024. [Online]. Available: \url{https://arxiv.org/abs/2312.11805}
\BIBentrySTDinterwordspacing

\bibitem{moo}
A.~Stone, T.~Xiao, Y.~Lu, K.~Gopalakrishnan, K.-H. Lee, Q.~Vuong, P.~Wohlhart, S.~Kirmani, B.~Zitkovich, F.~Xia, C.~Finn, and K.~Hausman, ``Open-world object manipulation using pre-trained vision-language models,'' 2023.

\bibitem{xiao2022dial}
T.~Xiao, H.~Chan, P.~Sermanet, A.~Wahid, A.~Brohan, K.~Hausman, S.~Levine, and J.~Tompson, ``Robotic skill acquistion via instruction augmentation with vision-language models,'' in \emph{Proceedings of Robotics: Science and Systems}, 2023.

\bibitem{myers2024palo}
\BIBentryALTinterwordspacing
V.~Myers, B.~C. Zheng, O.~Mees, S.~Levine, and K.~Fang, ``Policy adaptation via language optimization: Decomposing tasks for few-shot imitation,'' 2024. [Online]. Available: \url{https://arxiv.org/abs/2408.16228}
\BIBentrySTDinterwordspacing

\bibitem{zhang2024sprint}
J.~Zhang, K.~Pertsch, J.~Zhang, and J.~J. Lim, ``Sprint: Scalable policy pre-training via language instruction relabeling,'' in \emph{2024 IEEE International Conference on Robotics and Automation (ICRA)}.\hskip 1em plus 0.5em minus 0.4em\relax IEEE, 2024, pp. 9168--9175.

\bibitem{lynch2019relay}
A.~Gupta, V.~Kumar, C.~Lynch, S.~Levine, and K.~Hausman, ``Relay policy learning: Solving long horizon tasks via imitation and reinforcement learning,'' \emph{Conference on Robot Learning (CoRL)}, 2019.

\bibitem{susie}
\BIBentryALTinterwordspacing
K.~Black, M.~Nakamoto, P.~Atreya, H.~Walke, C.~Finn, A.~Kumar, and S.~Levine, ``Zero-shot robotic manipulation with pretrained image-editing diffusion models,'' \emph{ArXiv}, vol. abs/2310.10639, 2023. [Online]. Available: \url{https://api.semanticscholar.org/CorpusID:264172455}
\BIBentrySTDinterwordspacing

\bibitem{Fang2022PlanningTP}
\BIBentryALTinterwordspacing
K.~Fang, P.~Yin, A.~Nair, and S.~Levine, ``Planning to practice: Efficient online fine-tuning by composing goals in latent space,'' \emph{2022 IEEE/RSJ International Conference on Intelligent Robots and Systems (IROS)}, pp. 4076--4083, 2022. [Online]. Available: \url{https://api.semanticscholar.org/CorpusID:248834175}
\BIBentrySTDinterwordspacing

\bibitem{gu2023rttrajectory}
J.~Gu, S.~Kirmani, P.~Wohlhart, Y.~Lu, M.~G. Arenas, K.~Rao, W.~Yu, C.~Fu, K.~Gopalakrishnan, Z.~Xu, P.~Sundaresan, P.~Xu, H.~Su, K.~Hausman, C.~Finn, Q.~Vuong, and T.~Xiao, ``Rt-trajectory: Robotic task generalization via hindsight trajectory sketches,'' in \emph{Robotics: Science and Systems (RSS)}, 2024.

\bibitem{wen2023anypoint}
C.~Wen, X.~Lin, J.~So, K.~Chen, Q.~Dou, Y.~Gao, and P.~Abbeel, ``Any-point trajectory modeling for policy learning,'' \emph{Robotics: Science and Systems (RSS)}, 2024.

\bibitem{vima}
Y.~Jiang, A.~Gupta, Z.~Zhang, G.~Wang, Y.~Dou, Y.~Chen, L.~Fei-Fei, A.~Anandkumar, Y.~Zhu, and L.~Fan, ``Vima: General robot manipulation with multimodal prompts,'' in \emph{Fortieth International Conference on Machine Learning}, 2023.

\bibitem{nasiriany2024pivot}
S.~Nasiriany, F.~Xia, W.~Yu, T.~Xiao, J.~Liang, I.~Dasgupta, A.~Xie, D.~Driess, A.~Wahid, Z.~Xu \emph{et~al.}, ``Pivot: Iterative visual prompting elicits actionable knowledge for vlms,'' in \emph{Forty-first International Conference on Machine Learning}, 2024.

\bibitem{lee2019composing}
Y.~Lee, S.-H. Sun, S.~Somasundaram, E.~S. Hu, and J.~J. Lim, ``Composing complex skills by learning transition policies,'' in \emph{International Conference on Learning Representations}, 2019.

\bibitem{lee2019learning}
Y.~Lee, J.~Yang, and J.~J. Lim, ``Learning to coordinate manipulation skills via skill behavior diversification,'' in \emph{International conference on learning representations}, 2019.

\bibitem{shiarlis2018taco}
K.~Shiarlis, M.~Wulfmeier, S.~Salter, S.~Whiteson, and I.~Posner, ``Taco: Learning task decomposition via temporal alignment for control,'' in \emph{International Conference on Machine Learning}.\hskip 1em plus 0.5em minus 0.4em\relax PMLR, 2018, pp. 4654--4663.

\bibitem{gupta2019relay}
A.~Gupta, V.~Kumar, C.~Lynch, S.~Levine, and K.~Hausman, ``Relay policy learning: Solving long-horizon tasks via imitation and reinforcement learning,'' \emph{CoRL}, 2019.

\bibitem{shankar2019discovering}
T.~Shankar, S.~Tulsiani, L.~Pinto, and A.~Gupta, ``Discovering motor programs by recomposing demonstrations,'' in \emph{International Conference on Learning Representations}, 2019.

\bibitem{pertsch2020spirl}
K.~Pertsch, Y.~Lee, and J.~J. Lim, ``Accelerating reinforcement learning with learned skill priors,'' in \emph{Conference on Robot Learning (CoRL)}, 2020.

\bibitem{singh2020parrot}
A.~Singh, H.~Liu, G.~Zhou, A.~Yu, N.~Rhinehart, and S.~Levine, ``Parrot: Data-driven behavioral priors for reinforcement learning,'' \emph{ICLR}, 2021.

\bibitem{sharma2020learning}
M.~Sharma, J.~Liang, J.~Zhao, A.~LaGrassa, and O.~Kroemer, ``Learning to compose hierarchical object-centric controllers for robotic manipulation,'' \emph{arXiv preprint arXiv:2011.04627}, 2020.

\bibitem{dalal2021raps}
M.~Dalal, D.~Pathak, and R.~R. Salakhutdinov, ``Accelerating robotic reinforcement learning via parameterized action primitives,'' \emph{Neural Information Processing Systems (NeurIPS)}, vol.~34, pp. 21\,847--21\,859, 2021.

\bibitem{nasiriany2022maple}
S.~Nasiriany, H.~Liu, and Y.~Zhu, ``Augmenting reinforcement learning with behavior primitives for diverse manipulation tasks,'' in \emph{IEEE International Conference on Robotics and Automation (ICRA)}, 2022.

\bibitem{zhang2024extract}
J.~Zhang, M.~Heo, Z.~Liu, E.~Biyik, J.~J. Lim, Y.~Liu, and R.~Fakoor, ``Extract: Efficient policy learning by extracting transferrable robot skills from offline data,'' \emph{arXiv preprint arXiv:2406.17768}, 2024.

\bibitem{bacon2017option}
P.-L. Bacon, J.~Harb, and D.~Precup, ``The option-critic architecture,'' in \emph{Proceedings of the AAAI conference on artificial intelligence}, vol.~31, no.~1, 2017.

\bibitem{nachum2018data}
O.~Nachum, S.~S. Gu, H.~Lee, and S.~Levine, ``Data-efficient hierarchical reinforcement learning,'' \emph{Neural Information Processing Systems (NeurIPS)}, vol.~31, 2018.

\bibitem{peng2019mcp}
X.~B. Peng, M.~Chang, G.~Zhang, P.~Abbeel, and S.~Levine, ``Mcp: Learning composable hierarchical control with multiplicative compositional policies,'' \emph{Advances in Neural Information Processing Systems}, vol.~32, 2019.

\bibitem{hotspots}
T.~Nagarajan, C.~Feichtenhofer, and K.~Grauman, ``Grounded human-object interaction hotspots from video,'' in \emph{International Conference on Computer Vision (ICCV)}, 2019.

\bibitem{hoi}
S.~Liu, S.~Tripathi, S.~Majumdar, and X.~Wang, ``Joint hand motion and interaction hotspots prediction from egocentric videos,'' in \emph{{IEEE} Conference on Computer Vision and Pattern Recognition (CVPR)}, 2022.

\bibitem{vrb}
S.~Bahl, R.~Mendonca, L.~Chen, U.~Jain, and D.~Pathak, ``Affordances from human videos as a versatile representation for robotics,'' 2023.

\bibitem{swim}
R.~Mendonca, S.~Bahl, and D.~Pathak, ``Structured world models from human videos,'' 2023.

\bibitem{belkhale2024data}
S.~Belkhale, Y.~Cui, and D.~Sadigh, ``Data quality in imitation learning,'' \emph{Advances in Neural Information Processing Systems}, vol.~36, 2024.

\bibitem{morton2017simultaneouspolicylearninglatent}
\BIBentryALTinterwordspacing
J.~Morton and M.~J. Kochenderfer, ``Simultaneous policy learning and latent state inference for imitating driver behavior,'' 2017. [Online]. Available: \url{https://arxiv.org/abs/1704.05566}
\BIBentrySTDinterwordspacing

\bibitem{droidIL}
\BIBentryALTinterwordspacing
S.~Jayanthi, L.~Chen, N.~Balabanska, V.~Duong, E.~Scarlatescu, E.~Ameperosa, Z.~H. Zaidi, D.~Martin, T.~K.~D. Matto, M.~Ono, and M.~Gombolay, ``Droid: Learning from offline heterogeneous demonstrations via reward-policy distillation,'' in \emph{Proceedings of The 7th Conference on Robot Learning}, ser. Proceedings of Machine Learning Research, J.~Tan, M.~Toussaint, and K.~Darvish, Eds., vol. 229.\hskip 1em plus 0.5em minus 0.4em\relax PMLR, 06--09 Nov 2023, pp. 1547--1571. [Online]. Available: \url{https://proceedings.mlr.press/v229/jayanthi23a.html}
\BIBentrySTDinterwordspacing

\bibitem{hausman2017multimodalil}
\BIBentryALTinterwordspacing
K.~Hausman, Y.~Chebotar, S.~Schaal, G.~Sukhatme, and J.~Lim, ``Multi-modal imitation learning from unstructured demonstrations using generative adversarial nets,'' 2017. [Online]. Available: \url{https://arxiv.org/abs/1705.10479}
\BIBentrySTDinterwordspacing

\bibitem{arenas2023prompt}
M.~G. Arenas, T.~Xiao, S.~Singh, V.~Jain, A.~Z. Ren, Q.~Vuong, J.~Varley, A.~Herzog, I.~Leal, S.~Kirmani \emph{et~al.}, ``How to prompt your robot: A promptbook for manipulation skills with code as policies,'' in \emph{IEEE International Conference on Robotics and Automation (ICRA)}, 2024.

\bibitem{gemini15}
\BIBentryALTinterwordspacing
G.~Team, ``Gemini 1.5: Unlocking multimodal understanding across millions of tokens of context,'' 2024. [Online]. Available: \url{https://arxiv.org/abs/2403.05530}
\BIBentrySTDinterwordspacing

\bibitem{open_x_embodiment_rt_x_2023}
O.~X.~E. Collaboration \emph{et~al.}, ``Open x-embodiment: Robotic learning datasets and rt-x models : Open x-embodiment collaboration0,'' in \emph{2024 IEEE International Conference on Robotics and Automation (ICRA)}, 2024.

\bibitem{belkhale2024rt}
S.~Belkhale, T.~Ding, T.~Xiao, P.~Sermanet, Q.~Vuong, J.~Tompson, Y.~Chebotar, D.~Dwibedi, and D.~Sadigh, ``Rt-h: Action hierarchies using language,'' \emph{arXiv preprint arXiv:2403.01823}, 2024.

\end{thebibliography}
\bibliographystyle{IEEEtran}
\end{document}